\documentclass{article}
\usepackage{MLDD_workshop_2022,times}
\usepackage{microtype}
\usepackage{graphicx}
\usepackage{subfigure}
\usepackage{booktabs} 

\usepackage{url}
\usepackage{float}
\usepackage{graphicx}
\usepackage{babel,blindtext}

\usepackage{natbib}
\usepackage{hyperref}

\usepackage{amsmath}
\usepackage{amssymb}
\usepackage{mathtools}
\usepackage{amsthm}

\title{De novo design of protein target specific scaffold-based Inhibitors via Reinforcement Learning}

\author{Andrew D. McNaughton, Mridula S. Bontha, Carter R. Knutson, Jenna A. Pope, \\
\textbf{Neeraj Kumar}\thanks{Correspondence to \texttt{neeraj.kumar@pnnl.gov}} \\
Pacific Northwest National Laboratory\\
Richland, WA 99354, USA \\
}

\iclrfinalcopy 

\begin{document}

\maketitle


\begin{abstract}
Efficient design and discovery of target-driven molecules is a critical step in facilitating lead optimization in drug discovery. Current approaches to develop molecules for a target protein are intuition-driven, hampered by slow iterative design-test cycles due to computational challenges in utilizing 3D structural data, and ultimately limited by the expertise of the chemist – leading to bottlenecks in molecular design. In this contribution, we propose a novel framework, called 3D-MolGNN$_{RL}$, coupling reinforcement learning (RL) to a deep generative model based on 3D-Scaffold to generate target candidates specific to a protein building up atom by atom from the starting core scaffold. 3D-MolGNN$_{RL}$ provides an efficient way to optimize key features by multi-objective reward function within a protein pocket using parallel graph neural network models. The agent learns to build molecules in 3D space while optimizing the activity, binding affinity, potency, and synthetic accessibility of the candidates generated for infectious disease protein targets. Our approach can serve as an interpretable artificial intelligence (AI) tool for lead optimization with optimized activity, potency, and biophysical properties. 
\end{abstract}

\section{INTRODUCTION}

Recent advancements in machine learning (ML) and artificial intelligence (AI) have demonstrated the potential to revolutionize the drug design process by reducing the initial chemical search space in the early stages of discovery \citep{GCPN, born2020paccmann, Li2018GraphGen, huang2020deeppurpose, IBM}. Currently, the potential chemical space is composed of $>10^{60}$ molecules, candidates with suitable activity against specific protein targets only narrows the search space significantly based on the critical fragments. The COVID-19 pandemic has brought a surge of interest to explore data-driven methods to better produce efficacious drug candidates \citep{huang2020deeppurpose}. One of the most important factors in identifying new drug candidates is that the molecules possess optimized properties that allow them to effectively bind with a required target while also having minimal off-target effects. If we consider molecular generation as a controlled and dynamic step-by-step process, it is possible to produce end products that possess these optimized properties. This approach allows us to formulate \textit{de novo }drug design as a reinforcement learning (RL) problem and utilize algorithms that best learn a molecule's representation space based on the core moiety and its spatial interaction with the protein target. As an alternative, RL would provide a platform to create a highly-efficient inverse molecular design AI-system capable of producing novel high-performance molecules with domain-targeted properties.

\subsection{Related Works}

Our current work presents a novel approach to address the problem of generating  molecules optimized for specific 3D protein targets starting from core functional groups so called scaffold. In recent years, RL-based works to generate molecules have been introduced that attempt to tackle the problem in different ways. A method by \citet{Popova2018} utilizes a fragment-replacement-based approach to optimize existing SMILES strings for specific drug-like properties, while another method by \citet{Stahl2019} generates entirely new SMILES. Several current RL methods rely on the use of variational autoencoders (VAEs), which learn the latent space representation of 2D molecules to find the best representation \citep{Lim2018,Jin2018,LiuABG18,Sattarov2019,Joo2020}. However, none of these methods take into consideration the 3D protein structure during the generation phase. Some of these methods do test the drug-likeness post-generation against some given target protein, but the target is never directly involved in the RL loop. When designing drug candidates that target specific proteins, learning how the molecule interacts with the protein's structure is invaluable to the generative process and can greatly assist in accelerating the automation of medicinal chemistry. 

Recently, \citet{born2020paccmann} proposed a deep reinforcement learning framework called PaccMann$^{RL}$ for designing antiviral candidates binding against given protein targets using 2D protein sequence data. The generative model is composed of two pre-trained VAEs, namely a protein-VAE and a SELFIES-VAE, for mapping protein and drug molecules to a multi-modal latent space. The critic model is composed of two predictive models for predicting binding affinity of the protein-ligand pair and toxicity for the ligand. These predictions are used to formulate a multi-modal reward function to penalize the generative model \citep{born2021}. This method still uses VAEs to generate 2D molecules, but it combines the latent space of the SMILES with the embedding of a specific protein to generate SMILES that consider target protein information

\subsection{Problem Formulations and Proposed Method}

We propose a new method, known as 3D-MolGNN$_{RL}$, that not only incorporates the protein structure into the RL loop, but considers the 3D structures of the generated compounds by constructing them atom-by-atom in 3D space. To the best of our knowledge, this is a process that hasn't been explored as most methods simply rely on 2D representations of the molecules in the form of a SMILES string. To overcome the sequence-based representation of the target and the drug, we leverage a 3D-scaffold-based molecular generation actor over a 2D SMILES generation actor. The 3D-MolGNN$_{RL}$ method uses our previously formulated 3D-scaffold-based generative model \citet{joshi20213d} to make atom-wise placements on the molecule starting from a desired scaffold.

Our method considers a similar approach to that of the actor-critic method demonstrated in the PaccMann$^{RL}$ model. In our approach, we utilize a scaffold-based generative model, 3D-Scaffold \citep{joshi20213d} to produce valid 3D compounds as the actor model. For the critic model, we utilize parallel graph neural networks as a binding probability predictor (GNN$_{P}$ \citep{knutson2021decoding}) to evaluate whether the generated compound actively binds with a target protein. This method takes the 3D structure of the protein pocket and the ligand, and predicts the probability of their interaction without any prior knowledge of the intermolecular interactions. The GNN$_{P}$ model uses both the residue and 3D atomic-level representation of a protein as well as the 3D atomic and bond level representation of the molecule.

\section{EXPERIMENTS and DATASET }
\label{sec:Data}

To demonstrate our methods ability to produce molecules for a given target protein, this work was centered around creating drug-like molecules that interacted with the SARS-CoV-2 M\textsubscript{pro}. To train the model, We used a dataset of non-covalent inhibitors from the BindingDB dataset \citep{Gilson2016} and FDA approved drugs \cite{FDA}. These compounds were filtered based on biophysical and cheminformatics properties like IC\textsubscript{50}, molecular weight, atom type, and functionality and have almost 10K unique scaffolds. To represent the definition of scaffold, we used Murcko scaffolds \cite{Murcko}. A similar dataset was used in the prior 3D-Scaffold model \citep{joshi20213d}. High-throughput experimentation and our computer aided drug design results suggest that compounds containing the functional group piperazine are potent and have a higher affinity towards the M\textsubscript{pro} target \citep{clyde2021high, joshi20213d}. However, there were no piperazine-containing molecules in the initial dataset used for training the 3D-scaffold model, but we were still able to generate some molecules containing piperazine using the trained model. We curated a smaller subset of the BindingDB dataset that possessed affinity towards protease-like targets and combined them with the piperazine dataset. Finally, we filtered these compounds based on their ability to bind with M\textsubscript{pro} by doing an initial pass through the GNN$_{P}$ model. Since our motive is to achieve better binding affinity, potency and easily synthesizable compounds for the M\textsubscript{pro} target, the initial screening helped us to choose proper compounds for training the RL models so that they learned to produce similar, or better compounds.

\section{METHODS}

In this section, we discuss in detail the architecture and implementation of the RL approach. In particular,we go into detail about how the critic interprets the output from the actor.

\subsection{3D-MolGNN$_{RL}$ framework}

\begin{figure*}[htbp]
\vskip 0.2in
\begin{center}
\centerline{\includegraphics[width=100mm, scale=0.5]{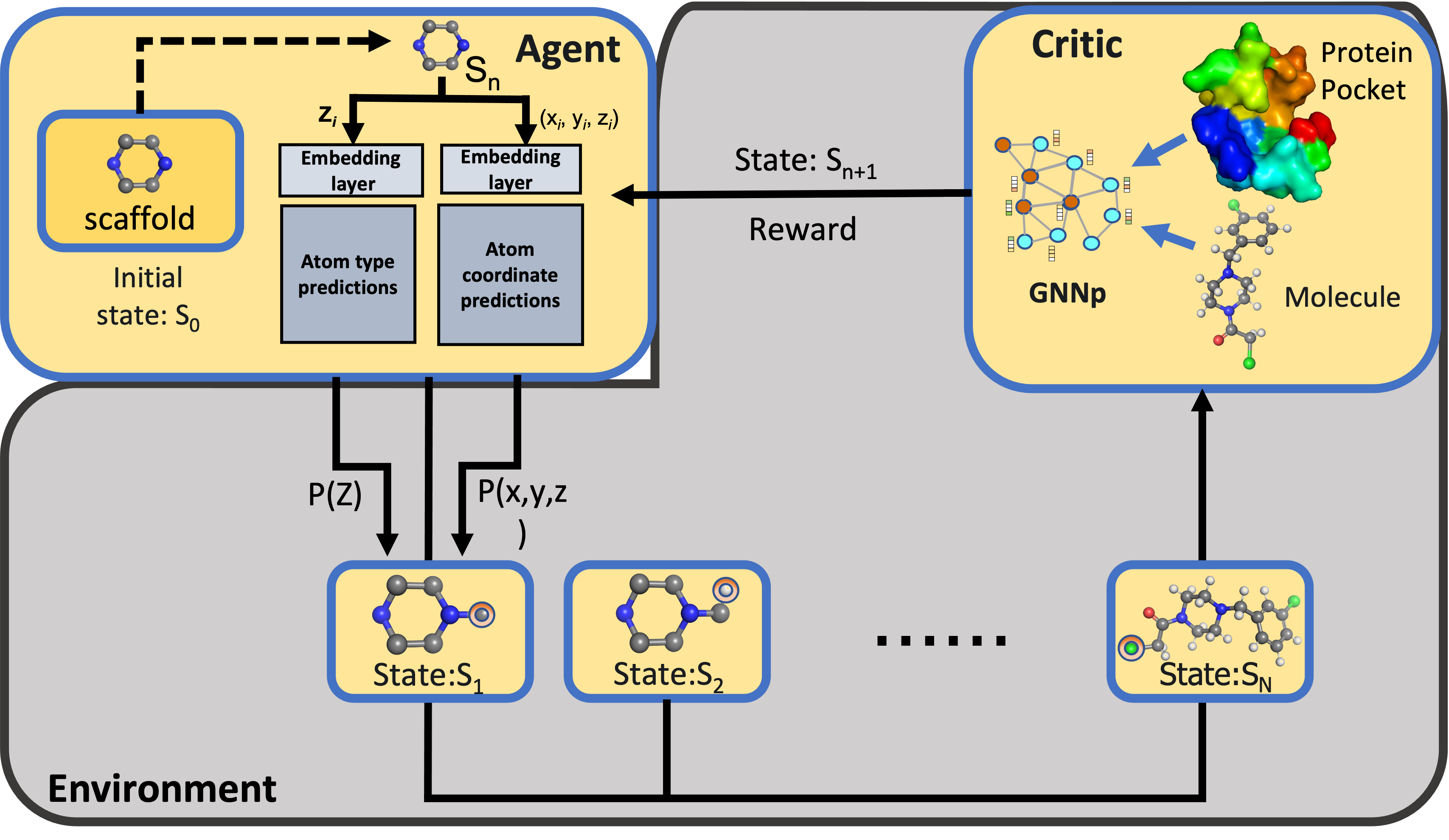}}
\caption{Schematic RL workflow highlighting the interaction of Actor and Critic in our 3D-MolGNN$_{RL}$ model. The agent starts building the molecule from a starting scaffold (state S\textsubscript{0}) and subsequently builds the molecule by choosing an atom based on the reward assigned by the critic for that intermediate state. The Agent block serves as the Actor, generating molecules to pass into the Critic block to assess the performance.}
\label{fig:3D-MolGNN}
\end{center}
\vskip -0.2in
\end{figure*}

The schematic representation of 3D-MolGNN$_{RL}$ framework is shown in the figure \ref{fig:3D-MolGNN}. The RL workflow begins from training a 3D-scaffold molecule generator followed by optimizing the partially built candidates towards the target of interest at every step of training. 
The neural network used in the 3D-Scaffold framework for \textit{de-novo} drug candidate design can be broken into two major blocks: feature learning and atom placement as shown in Figure \ref{fig:3D-MolGNN}. In the feature learning block, the embedding and interaction layers of SchNet \citep{joshi20213d, schnet1, schnet2, schnet3, schnet4} are used to extract and update rotationally and translationally invariant atom-wise features that capture the chemical environment of an unfinished molecule. The extracted features are used to predict distributions for the type of next atom and its 3D coordinates, where the latter distribution is constructed from predictions of pairwise distances between the next atom and all preceding atoms. The whole procedure is repeated successively to build a complete molecule with the desired scaffold. The partial molecule associated with each step of this atom placement is assessed by the critic (GNN$_{P}$). As a part of our experimentation, we used two critics simultaneously to use the binding probability and binding affinity of the partial molecule and the target of interest along with the synthetic accessibility of the partial molecule. The binding probability is the outcome from the SoftMax layer which predicts the activity/inactivity of a protein-ligand complex. The binding affinity is measured in terms of K\textsubscript{i} and K\textsubscript{d}, which refer to the inhibition and dissociation constants, respectively and is given as $-log(\frac{K_{i}}{K_{d}})$.The two critics essentially use the same feature representation, but differ in terms of the data they have been trained on and also the label associated with the data. A more detailed breakdown of the layers and hyperparameters associated with the actor and critic models is included in the Appendix.

\subsection{Reward Function \& Parameters}

The process of building the molecule starts from a desired scaffold associated with the selected core functionality. For every action \textit{t} which is a random selection and placement of the atom to the partial molecule, the 3D-scaffold model predicts the next possible atom that could be placed close to the center of mass to the partial molecule at step $t-1$ and transitions to state $s_t$. At any step, let $Z_{\text{next}}$ be the ground truth type of the next atom and $\hat{p}_{\text{type}}^{Z_{\text{next}}}$ the probability that the model assigns to that type at the current step. This probability is converted to a negative log-likelihood as $P_\Theta(s_t)^{type}$ (equation \ref{eq:3a}). 
Here, $\hat{p}_j^b$ is the probability that the model assigns for the distance between position of already placed atom and ground truth of the next atom to fall into distance bin $b\in B$ at the current step. The distance based probability is calculated using Gaussian expanded ground truth distances $q_j^b$ and probability distribution of atom placement $\hat{p}_j^b$ s shown in equation \ref{eq:3b}.
The overall idea is to train the agent to learn the latent space representation of atom type and it's possible placement in 3D-space closest to the center of mass of the molecule generated so far. The overall probability for any action \textit{t} is the summation of type-probability and distance-probability (equation \ref{eq:2}). The ultimate goal is to train the agent such that it learns to generate new compounds while optimizing the policy $\Pi(\Theta)$ (equation \ref{eq:1}). The policy is defined as the difference between the action-probabilities and the reward assigned by the critic at that step. 

\begin{equation}\label{eq:1}
\Pi(\Theta) = \sum_{s_{t} \in S} (P_\Theta(s_t) - R(s_t)) 
\end{equation}%
\begin{equation}\label{eq:2}
P_\Theta(s_t) = P_\Theta(s_t)^{type} + P_\Theta(s_t)^{dist}
\end{equation}%
\begin{subequations}\label{eq:3}
\begin{minipage}{0.45\textwidth}
\begin{align}
\label{eq:3a}
P_\Theta(s_t)^{type} = -\log \left(\hat{p}_{\text{type}}^{Z_{\text{next}}}\right)
\end{align}
\end{minipage}
\begin{minipage}{0.45\textwidth}
\begin{align}
\label{eq:3b}
P_\Theta(s_t)^{dist} = \sum_{j=1}^{N}\sum_{b\in B} q_j^b \log
\left(\hat{p}_j^{b}\right) 
\end{align}
\end{minipage}
\end{subequations}
\begin{equation}\label{eq:5}
R_1(s_t) = \alpha.C_{BP}(B,C_t) + \beta.C_{EA}(B,C_t) + (1 -\gamma.C_{SA}(C_t))
\end{equation}
\begin{equation}\label{eq:4}
R_2(s_t) = \alpha.C_{BP}(B,C_t) + (1-\beta.C_{SA}(C_t))
\end{equation}%

We used two different reward functions as $R_1(s_t)$ and $R_2(s_t)$ associating with two independent experiments for the purposes of ablation/hyperparameter testing. By removing portions of the reward function, we can see which attributes provide the best reward. The first reward is a function of binding probability, binding affinity of the target and the partially built molecule in addition to the synthetic accessibility of the molecule. While the second reward function uses only the binding probability and the synthetic accessibility score. We assigned different weights to each of these components while calculating the rewards. For $R_1(s_t)$ (equation \ref{eq:4}), we used 0.5,0.25 and 0.25 respectively for binding probability, binding affinity and synthetic accessibility. While for $R_2(s_t)$ (equation \ref{eq:5}, we used 0.75 and 0.25 for binding probability and synthetic accessibility respectively. We trained the agent for 150 epochs and chose the best trained model to generate compounds.The Synthetic Accessibility and Binding Affinity scores are scaled to be in between 0 and 1 to match the scale of the binding probability. For more on the ablation study, including some reward functions not shown in the main text, see Appendix \ref{app:ablation}.

\section{RESULTS \& DISCUSSION}

\subsection{Model Performance}

\begin{figure}[h]
    \centering
    \includegraphics[scale=0.5]{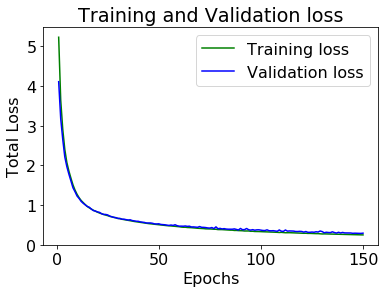}
    \caption{Plot of loss over total epochs, the loss represents how well the model learns the 3D trace of the input data. Loss is calculated as a combination of KL-Divergence and the Reward Function}
    \label{fig:total_loss}
\end{figure}

To ensure that we are a learning to build molecules based on the training data, we visualize the training and validation error over each epoch. Figure \ref{fig:total_loss} represents the aggregated loss over the 150 epochs for our 3D-MolGNN$_{RL}$ model. As loss decreases over time, we know that our model is able to effectively learn from the training data. This result is strictly evaluating the training and validation loss through the actor model to evaluate training performance, the drug-likeness scoring of the GNN model is discussed further in the next section.

\subsection{Drug-likeness Metrics for Generated Compounds}

We examined the effectiveness of our RL methods using different drug-likeliness metrics: quantitative estimate  of drug-likeness (QED), water solubility (logS), synthetic accessability (SA), and the octanol-water partition coefficient (logP). See the appendix for a full explanation of each metric. A desirable drug candidate would score well in each of these metrics. Since the agent for the 3D-MolGNN$_{RL}$ is a scaffold based generative model, we focused on generating compounds with piperazine as the scaffold. 
To ensure that only valid molecules are compared, we filtered the total list of generated compounds based on a modified Lipinksi rule. The rule suggests that for a candidate to be an acceptable drug-like compound, there should be no more than 5 hydrogen bond donors, no more than 10 hydrogen bond acceptors, no more than 5 rotatable bonds, a molecular mass between 200 and 500 Dalton, an octanol-water partition coefficient less than 5, and finally at least one aromatic ring in the structure. Once filtered, approximately 100 top compounds per method were obtained.

\subsubsection{Comparison to 2D Generative Model}

To show the advantages of using 3D-based generative modeling over conventional 2D-based generative models (SMILES-based), we compare our 3D-MolGNN$_{RL}$ model to that of PaccMann$_{RL}$. For better comparison of the two methods, we utilize a slightly modified version of the architecture to include our GNN critic model instead of the default PaccMann critic. This allows us to have a metric of comparison between a 2D and 3D generative model. For purposes of discussion, the modified PaccMann$_{RL}$ model will be referred to as PM-GNN$_{RL}$.

\begin{figure}[!ht]
    \centering
    \subfigure[\centering 3D-MolGNN-RL optimized vs unoptimized]{\includegraphics[width=6cm]{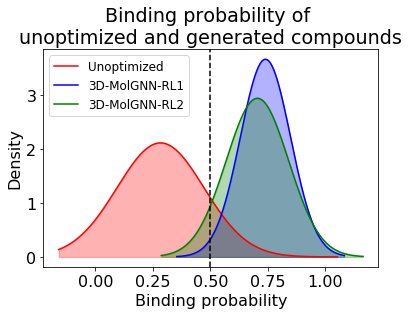}}
    \subfigure[\centering PM-GNN-RL optimized vs unoptimized]{\includegraphics[width=6cm]{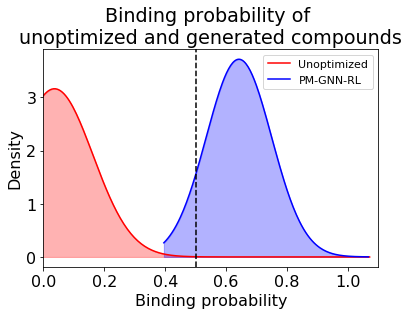}}
    \caption{Comparison of generated compounds before and after optimization by 3D-MolGNN$_{RL}$. The unoptimized molecules were generated using the respective agents for each method over the same data but without involvement of the critic. Results visualized as the normalized probability density function of the data}
    \label{fig:unoptimized_vs_biased}
\end{figure}

We first wanted to demonstrate that the RL implementation of the methods were essential in producing molecules with a high binding probability. To accomplish this, we generated molecules for the protein target M\textsubscript{pro} (PDB:6WQF), both with and without passing the reward from the critic back to the agent in each method. We designated here the molecules generated without RL optimization as \textit{unoptimized}. To fairly compare the optimized vs unoptimized methods, we ensured that the unoptimized models were trained and tested on the same datasets used for training the respective RL models. The comparison of binding probabilities of the generated compounds towards the M\textsubscript{pro} target from the 3D-MolGNN$_{RL}$ and PM-GNN$_{RL}$ is shown in figure \ref{fig:unoptimized_vs_biased}a and \ref{fig:unoptimized_vs_biased}b, respectively. From this figure, it is evident that the RL mechanism has improved both agent's performance in producing candidates with high binding probability. The unoptimized molecules generated by the agent in 3D-MolGNN$_{RL}$ and PM-GNN$_{RL}$ show very low binding probability towards M\textsubscript{pro} as the predictions have a low mean probability. On the other hand, when using biased compounds during RL optimization, we see a significant increase in high binding probability compounds, showing the effects of incorporating RL with the generative models.

\begin{figure}[ht]
    \centering
    \subfigure[\centering Drug-likeness models]{{\includegraphics[width=6cm]{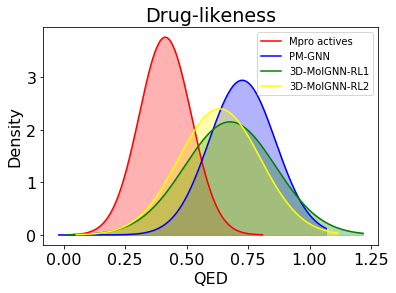}}}%
    \qquad
    \subfigure[\centering ESOL models]{{\includegraphics[width=6cm]{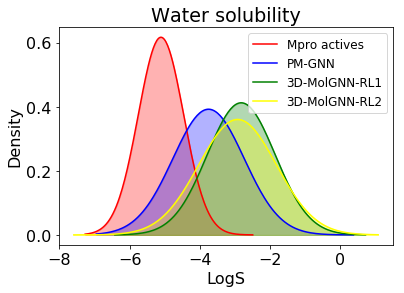}}}\\
     \subfigure[\centering Synthetic accessibility score models]{{\includegraphics[width=6cm]{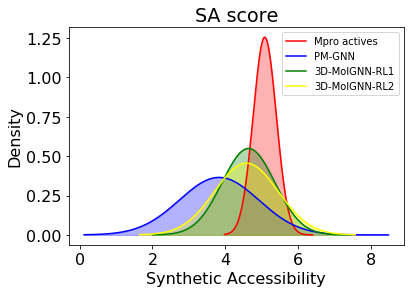}}}%
    \qquad
    \subfigure[\centering Partition coefficient models]{{\includegraphics[width=6cm]{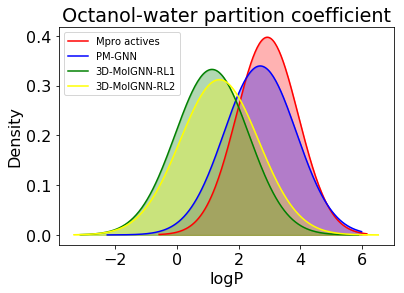}}}\\
    \caption{Comparison of the properties of the molecules produced from 3D-MolGNN$_{RL}$ and PM-GNN$_{RL}$ against the experimentally identified active compounds for M\textsubscript{pro} target. Results visualized as the normalized probability density function of the data}%
    \label{fig:metric}%
\end{figure}

To further assess the novelty and properties of generated compounds, we looked at the four different drug-likeness properties listed above to serve as comparison metrics. Figure \ref{fig:metric} compares these metrics for both 3D-MolGNN$_{RL}$ reward functions with known active molecules for M\textsubscript{pro}. A full description of each metric is given in the appendix (section \ref{sec:per_met}). 
The first metric, QED, represents a quantification of the desirability of the drug \citep{Bickerton2012}. We can see that in Figure \ref{fig:metric}a, the RL methods all produce higher scoring molecules than the current experimentally determined M\textsubscript{pro} actives with the mean of the PM-GNN$_{RL}$ having a slightly higher increase in mean.

We next look at the distributions of water solubility calculated using the ESOL method \citep{Delaney2004}. Figure \ref{fig:metric}b shows that our methods score better than the current M\textsubscript{pro} actives while also showing that these methods produce molecules with desirable solubility in water. The mean  of the 3D-MolGNN$_{RL}$'s $R_1(s,t)$ model has the highest improvement relative to the mean of the M\textsubscript{pro} actives while both the $R_2(s,t)$ and PM-GNN$_{RL}$ models improve less.

The next metric, synthetic accessibility (SA), is a term included in each of the reward functions of the RL methods tested. Figure \ref{fig:metric}c illustrates the distributions obtained by each method where a SA score closer to 1 is preferred. The two 3D-MolGNN$_{RL}$ models have a consistent range of SA scores that center around 5, which indicates consistently above average scores. $R_1(s,t)$ shows an 8.78\% improvement and $R_2(s,t)$ shows a 9.92\% improvement. The PM-GNN$_{RL}$ spans a much wider range of SA scores, but the mean is centered around 4. These three models only slightly edge out the existing M\textsubscript{pro} actives.

The last metric is the octanol-water partition coefficient, or more simply known as log$P$. Looking at Figure \ref{fig:metric}d, we can see how each method produces molecules spanning the range of -2 to 6. The 3D-MolGNN$_{RL}$ model using the $R_1(s,t)$ reward yields a 61.21\% improvement to the M\textsubscript{pro} actives in logP value, whereas using the $R_2(s,t)$ reward yields a 53.08\% improvement. In this metric, the PM-GNN$_{RL}$ metric produces compounds very similar to the M\textsubscript{pro} actives.

\begin{table*}[htbp!]
\centering
\caption{Details about the drug-likeness metrics proposed in this work. We look at the 3D-MolGNN$_{RL}$ model's two separate reward functions $R_1(s,t)$ and $R_2(s,t)$. For each method, the top 3 candidates are chosen and their metric scores listed.}
\tabcolsep=0.12cm
\begin{tabular}{lllllllllllll}
\hline
             & \multicolumn{3}{l}{\textbf{QED}} & \multicolumn{3}{l}{\textbf{logS}} & \multicolumn{3}{l}{\textbf{SA}} & \multicolumn{3}{l}{\textbf{logP}} \\
Rank &
  1st &
  2nd &
  3rd &
  1st &
  2nd &
  3rd &
  1st &
  2nd &
  3rd &
  1st &
  2nd &
  3rd \\ \hline
3D-MolGNN$_{RL}$ R1 & 0.49   & 0.59   & 0.41   & -3.32   & -2.49  & -2.23  & 4.81   & 4.81   & 4.59  & 1.56    & 0.82   & -0.14   \\ 
3D-MolGNN$_{RL}$ R2 & 0.32   & 0.44   & 0.82   & -2.53   & -3.82  & -2.90  & 3.55   & 4.84   & 3.93  & 0.71    & 2.29   & 1.68   \\ 
PM-GNN$_{RL}$       & 0.45   & 0.39   & 0.74   & -3.14   & -3.90  & -3.31  & 2.86   & 2.22   & 2.27  & 1.53    & 2.70   & 1.78   \\ \hline 
\end{tabular}
\label{table:metrics}
\end{table*}

To compare how each method performed on a per-molecule basis, Table \ref{table:metrics} gives the results for the top 3 candidates produced by each of the methods. These molecules were ranked based on their predicted binding probability with the M\textsubscript{pro} target. We can see that molecules that are performing above average in one metric are not guaranteed to perform well in every metric. For example, the top candidates for each method are each above the mean SA for the given method, but produce below the mean for QED. 2D and 3D snapshots of the top 3 candidates from each method are available in appendix section \ref{sec:top}.

Since our goal is achieve \textit{de novo} design, additional metrics such as validity, uniqueness, and novelty were considered among the piperazine-based generated compounds as shown in Table \ref{table:percentages}. Initial validity was determined by processing the molecules using cheminformatics tools. The methods of calculating these properties are described in further detail in appendix subsection \ref{sec:vun}. From table \ref{table:percentages}, we found that 3D-MolGNN$_{RL}$ R1 and 3D-MolGNN$_{RL}$ R2 produced greater than 95\% valid compounds while also scoring close to 100\% in uniqueness and novely from this valid set. The PM-GNN$_{RL}$ model, while not scoring quite as well in Uniqueness and Novelty, scores the highest in Validity. This shows that by incorporating more parameters into the multi-objective reward function, there is an improvement in the generation of novel drug candidates. Overall, 3D-MolGNN$_{RL}$ R1 outperformed other methods by generating compounds with high Validity, Uniqueness and Novelty.

\begin{table}[h]
\caption{Table outlining the three metrics used to evaluate the compounds produced by varying reward functions and RL models. Compounds were screened and scored for an overall percentage based on validity, uniqueness, and novelty.}
\tabcolsep=0.14cm
\begin{center}
 \begin{tabular}{llll}
\hline
Model & Validity & Uniqueness & Novelty \\ \hline
3D-MolGNN$_{RL}$ R1 & 99.9\% & 100\% & 99.9\% \\
3D-MolGNN$_{RL}$ R2 & 97\% & 99.9\% & 99.9\% \\
PM-GNN$_{RL}$ & 100\% & 93\% & 93\%  \\ \hline 
\end{tabular}
\label{table:percentages}   
\end{center}
\end{table}

\section{CONCLUSIONS}

In this work, we introduced a new method to include both the 3D structure of protein target and the generated compounds to perform multi-objective lead optimization critical for drug design and discovery. We demonstrated that our novel framework 3D-MolGNN$_{RL}$, which couples RL to a deep generative model based on a 3D-Scaffold, can generate target candidates built up atom by atom that are specific to a protein pocket. 3D-MolGNN$_{RL}$ provides an efficient way to generate target specific candidates by learning to build molecules in 3D space while optimizing the binding affinity, potency, and synthetic accessibility. To accomplish this, we utilized the protein for SARS-CoV-2 M\textsubscript{pro} as a target for generating optimized inhibitor candidates. We found that our model was able to generate molecules with better druglikeness, synthetic accessibility, water solubility, and hyrdophilicity than current M\textsubscript{pro} active molecules. This was given by a $>$50\% increase in QED, a $>$40\% increase in solubility, a $>$8\% improvement in SA, and a $>$50\% improvement in hydrophilicity. We found that our RL integration significantly improved the types of molecules generated by the untrained agents. Throughout this work, we demonstrated that by including more parameters into the multi-objective reward function, there is an improvement in generated novel target specific candidates.  This gives us confidence that our RL framework is effective at producing protein target specific hit candidates by leveraging the 3D structures of both the generated molecule and the protein pocket, a consideration not made by other molecular generation methods to date.

\section*{Acknowledgements}
This work was supported in part by the U.S. Department of Energy, Office of Science, Laboratory Directed Research Funding (LDRD), Mathematics of Artificial Reasoning for Science (MARS) Initiative, at the Pacific Northwest National Laboratory. Pacific Northwest National Laboratory (PNNL) is a multiprogram national laboratory operated by Battelle for the DOE under Contract DE-AC05-76RLO 1830. This research used computational resources provided by Research Computing at the Pacific Northwest National Laboratory.

\bibliographystyle{MLDD}
\bibliography{ms.bib}

\newpage
\appendix
\onecolumn
\section{Performance Metrics}
\label{sec:per_met}

\subsection{Quantitative Estimate of Druglikeness}

The QED metric represents a quantification of the desirability of the drug \citep{Bickerton2012}. The closer the score is to 1, the more desirable it is as a drug candidate.

The equation for QED is given as:

\[QED = \exp\left(\frac{1}{n}\sum_{i=1}^{n}\ln{d_{i}}\right),\]
\label{eq:QED}

Where $d_i$ is a series of desirability functions (d) belonging to eight widely used molecular descriptors. These are molecular weight (MW), the octanol-water partition coefficient (ALOGP), the number of hydrogen bond donors (HBD), the number of hydrogen bond acceptors (HBA), the molecular polar surface area (PSA), the number of rotatable bonds (ROTB), the number of aromatic rings (AROM), and the number of structural alerts (ALERTS).

The desirability function can be represented by a general asymmetric double sigmoidal function where d(x) is the desirability function for molecular descriptor x shown as:

\[d_i(x)=a_i+\frac{b_i}{1+\exp\left({-\frac{x-c_i+\frac{d_i}{2}}{e_i}}\right)}\cdot\left[ 1- \frac{1}{1+\exp\left({-\frac{x-c_i+\frac{d_i}{2}}{e_i}}\right)}
\right]
,\]

where $a_i,...,f_i$ can be found in the supplementary table of the original publication \citep{Bickerton2012}.

\subsection{Estimating Aqueous Solubility}

The metric, water solubility, calculates the log solubility (log$S$) of the molecule. In this work, the solubility is determined by ESOL \citep{Delaney2004}. The majority of drugs posses a log$S$ between -8 and -2. The more positive the value, the more water soluble the molecule.

ESOL as defined in \citet{Delaney2004} can be calculate as the multiple linear regression of:
\begin{enumerate}
\item{clogP}
\item{Molecular weight (MWT)}
\item{Rotatable bonds (RB)}
\item{Aromatic proportion (AP)}
\end{enumerate}
given as:

\[Log(S_w) = 0.16 - 0.63clogP - 0.0062MWT + 0.066RB - 0.74AP
\]

\subsection{Synthetic Accessibility}

SA is one of the most critical metrics to use in determining the simplicity in experimentally synthesizing a molecule. It is not a score that dictates how effective the molecule is, but rather a practical measure of it's complexity. The SA score is between 1 to 10, where 1 indicates an easily synthesizable molecule and 10 indicates a complex one.

The algorithm for calculating the SA score of a molecule (as represented in \citet{Ertl2009}) is given as:

\[SAscore = fragmentScore - complexityPenalty ,
\]
where the fragment score is calculated as a sum of contributions of all fragments in the molecule divided by the number of fragments in this molecule. The contribution of a fragment is obtained from a database of fragment contributions that were generated by statistical analysis of substructures in the PubChem collection. 

The complexity penalty is a score given to characterize the presence of complex structural features in the molecules. It is defined as a combination of the following:

\[ringComplexityScore = log(nRingBridgeAtoms +1) + log(nSpiroAtoms + 1)\]
\[stereoComplexityScore = log(nStereoCenters + 1)\]
\[macrocyclePenalty = log(nMacrocycles + 1)\]
\[sizePenalty = nAtoms^{1.005} - natoms\]

\subsection{Hydrophilicity vs. Lipophilicity}

This metric, known as log$P$, is the calculated octanol-water partition coefficient of a given molecule. The values represents if a drug is either very hydrophilic (-3) or very lipophilic (+10). This specific metric is present in Lipinski's rule as value that needs to be less than 5 to be considered a drug candidate.

To calculate the partition function for octanol-water partition coefficient that dictates whether a molecule is more hydrophilic or lipophilic we utilize the RDKit package implementation of the Crippen approach \citep{Wildman1999}. It simply calculates the sum of the contributions of each of the atoms in the molecule. Intramolecular interactions are accounted for my classifying atoms into different types based on their attached $a_i$ and neighboring atoms $n_i$ :

\[P_{calc} = \sum_{i}n_ia_i
,\]
where P can be further calculated into $logP$. A full list of the atomic descriptors and contributions can be found in the main text of \citet{Wildman1999}.

\subsection{Validity, Uniqueness, and Novelty}
\label{sec:vun}
To analyze the novelty of our compounds, we need to look at how we calculate the validity and uniqueness of our compounds.

These are given as follows:

\[ \text{Validity} = \frac{\text{Number of valid molecules}}{\text{Number of generated molecules}}
,\]

\[ \text{Unique} = \frac{\text{Number of unique molecules}}{\text{Number of valid molecules}}
,\]

\[ \text{Novelty} = \frac{\text{Number of generated molecules not in training set}}{\text{Number of unique and valid generated molecules}}
\]
\newpage

\section{Top Candidates}
\label{sec:top}

Shown below are snapshots of the top 3 candidates from each of the reward functions presented in this work.

\subsection{2D Representations}

\begin{figure}[h]

    \subfigure[3D-MolGNN$_{RL}$ - R2 Top 3 candidates: 94.49\%, 91.02\%, 84.88\%]{{\includegraphics[width=\textwidth]{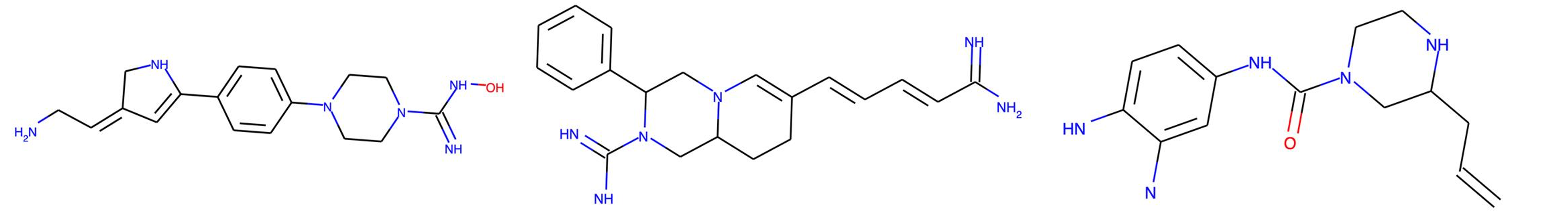}}}\\
    \subfigure[\centering 3D-MolGNN$_{RL}$ - R1 Top 3 candidates: 91.31\%, 91.82\%, 88.07\%]{{\includegraphics[width=\textwidth]{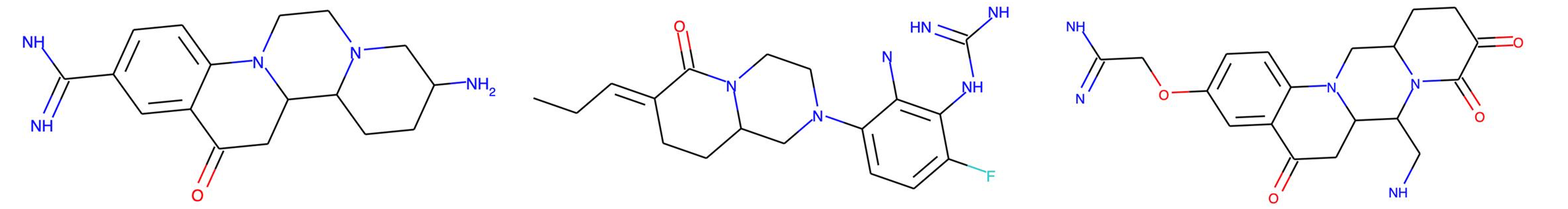}}}\\
    \caption{The top 3 candidates for each reward function in this work. The candidates are shown in descending order from left to right. Their associated binding probability is listed in the subcaption also in descending order from left to right.}%
    \label{fig:top3}%
\end{figure}
\newpage
\subsection{3D Representations}

\begin{figure}[h!]

    \subfigure[\centering 3D-MolGNN$_{RL}$ - R1 Top 3 candidates: 91.31\%, 91.82\%, 88.07\%]{\includegraphics[width=\textwidth]{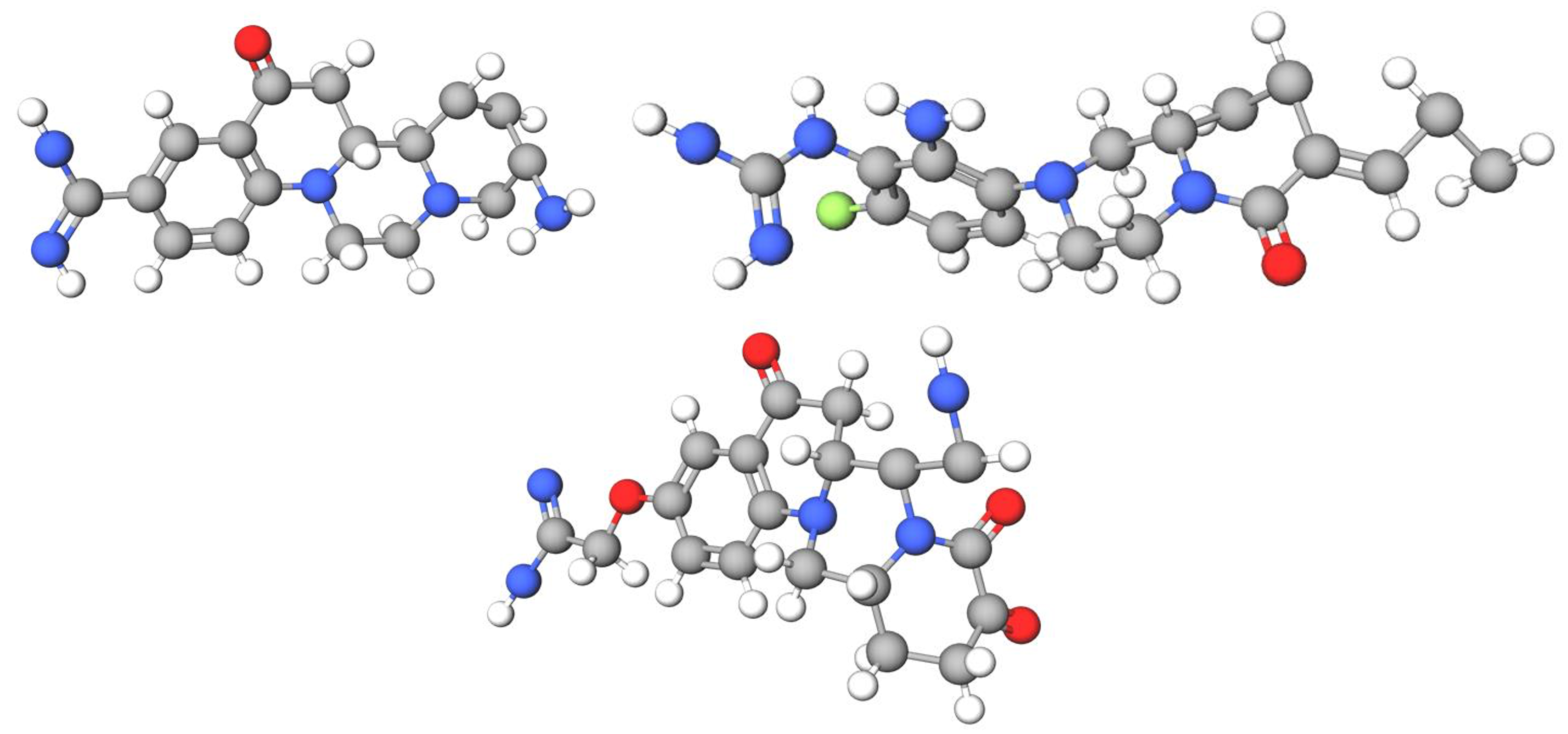}}\\
    \subfigure[\centering 3D-MolGNN$_{RL}$ - R2 Top 3 candidates: 94.49\%, 91.02\%, 84.88\%]{{\includegraphics[width=\textwidth]{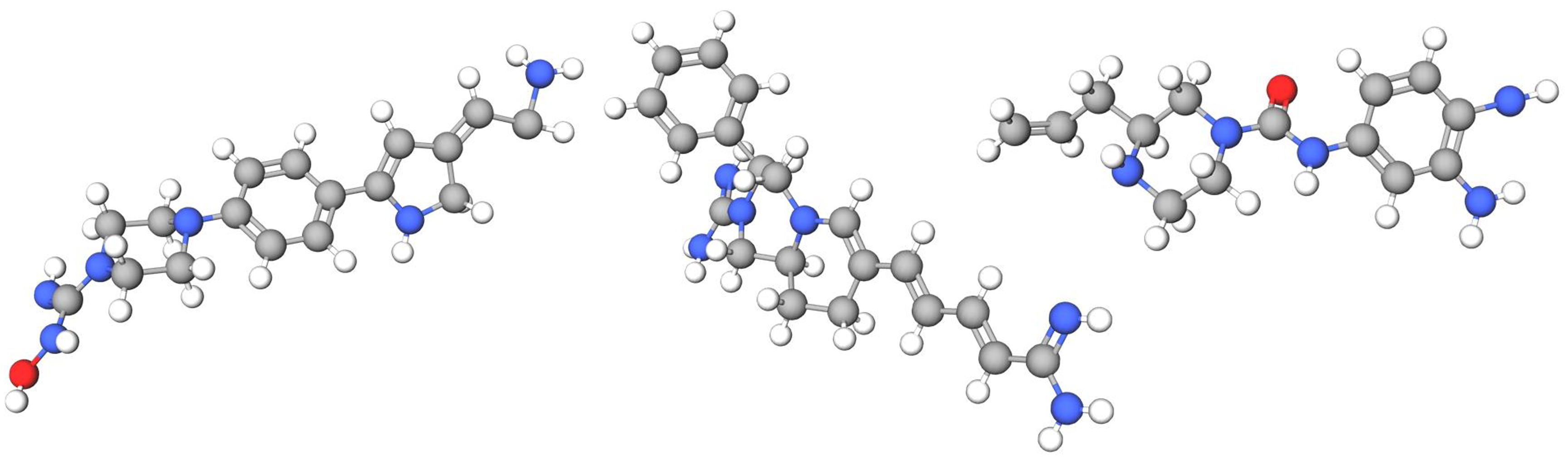}}} \\
    \caption{The top 3 candidates for each reward function in this work. The candidates are shown in descending score order from left to right and down. Their associated binding probability is listed in the subcaption also in descending order from left to right.}%
    \label{fig:3D_top3}%
\end{figure}
\newpage

\section{Ablation Study on Reward Function}
\label{app:ablation}

In order to see the importance of optimizing for multiple properties in the reward function, we tested different training weight, and reward function combinations for the model. The most important 2 were listed in the main text of the paper, but we tested 2 more reward functions to test performance.

We used the following combinations and compared how their output scored in the drug-like metrics. We see a steady decrease in performance from Reward 1 to Reward 3 showing that the inclusion of Binding Probability is important for generating molecules with a drug-like features. We also compare our 3D method with the 2D protein inclusion using the original PaccMann Method but with our GNN instead of their PaccMann critic to see if the 2D features have an effect.

$R_1(s_t) = \alpha.C_{BP}(B,C_t) + \beta.C_{EA}(B,C_t) + (1 - \gamma.C_{SA}(C_t))$

$R_2(s_t) = \alpha.C_{BP}(B,C_t) + (1 - \beta.C_{SA}(C_t))$

$R_3(s_t) = \alpha.C_{EA}(B,C_t) + (1 - \beta.C_{SA}(C_t))$

\begin{figure}[!htp]
    \centering
    \subfigure[\centering Drug-likeness models]{{\includegraphics[width=6cm]{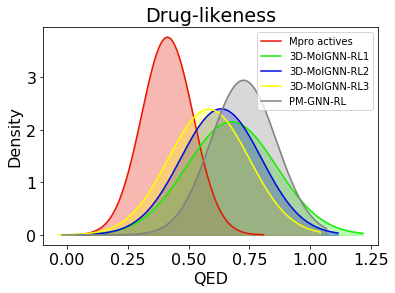}}}%
    \qquad
    \subfigure[\centering ESOL models]{{\includegraphics[width=6cm]{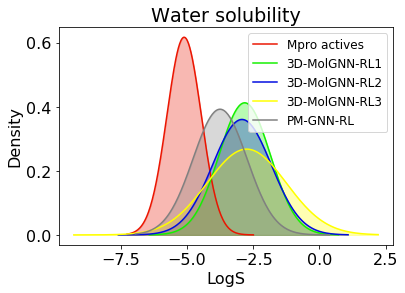}}}\\
     \subfigure[\centering Synthetic accessibility score models]{{\includegraphics[width=6cm]{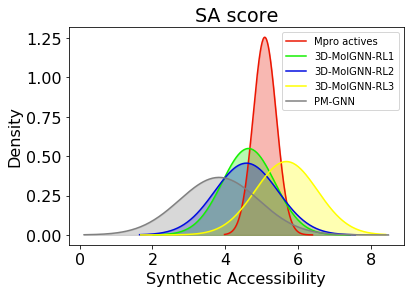}}}%
    \qquad
    \subfigure[\centering Partition coefficient models]{{\includegraphics[width=6cm]{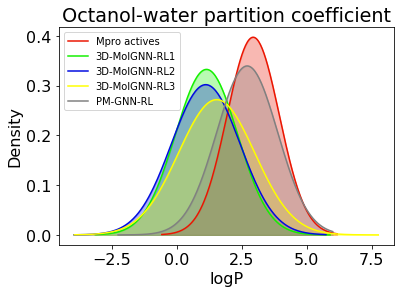}}}\\
    \caption{Comparison of the properties of the molecules produced from 3D-MolGNN$_{RL}$ against the experimentally identified active compounds for M\textsubscript{pro} target.}%
    \label{fig:ab_metric}%
\end{figure}

Overall, figure \ref{fig:ab_metric} shows that the complete 3 term reward function for the 3D-MolGNN$_{RL}$ method produces the best scoring molecules overall for the given drug-like metrics.  

\section{3D-Scaffold Hyperparameters}

For the 3D Scaffold model we use the following hyperparameters: batch size of 2, split of 1000-500 training, validation set, 150 max epochs, learning rate of 0.0001, learning rate patience of 10, learning rate minimum of 1e-6, learning rate decay of 0.5, atom-wise representation layer size of 32, cutoff radius of local environment of 10, interaction layer size of 6, 25 gaussians to expand distances with a max distance of 15 and 300 distance bins.

\section{GNN Hyperparameters}
Hyperparameters such as network depth, layer dimension, and learning rate can have a large effect on model training and the weights in the final realized model. Therefore, we performed a number of trainings to examine combinations of learning rate, number of attention heads, and layer dimension.  We used the parameters from our best model to create a baseline: a learning rate of 0.0001, four attention heads, and a layer dimension of 70 produced an average test AUROC of 0.854. The test parameters and top scoring configuration are outlined and highlighted in Table 3. The hyperparameters that performed best were a learning rate of 0.0001, two attention heads, and a dimension of 70. These parameters resulted in an average test AUROC of 0.864, still scoring lower than our implemented baseline configuration.

\begin{table}[h]
\centering
\caption{Values of examined hyperparameters: learning rate (lr), number ($N$) of attention heads, and dimension ($D$) of the GAT layer in each attention head. A grid search was performed on each combination of parameters. The optimal combination is shown in bold.}
\label{tab:hypTable_SI}
\begin{tabular}{lcc}
    \hline
lr & $N$ & $D$  \\
    \hline
0.001 & 2 & \textbf{70} \\
\textbf{0.0001} & 3 & 140 \\
0.00001 & \textbf{4} & 210 \\
 & & 280 \\
    \hline
  \end{tabular}
\end{table}

Trainings were carried out over 200 epochs on a quarter of the data taken from our dataset consisting of 2,000 samples per target with 79 targets, using the same train-test split for all trainings. The decreased performance compared to the models is expected due to the reduced set of data used to train the model.
Thirty of the 36 combinations trained without error and are shown in Table 4.

\begin{table}[H]
    \caption{Values of examined hyperparameter set with average train and Test ROC .}
    \label{tab:hypTableSI}
\vskip 0.15in 
\begin{center}
\begin{sc}
\begin{tabular}{lcc}
\toprule
HYPERPARAMETER SET & TRAIN ROC AVG & TEST ROC AVG \\
\midrule
Lr\textunderscore 0.001\textunderscore n5\textunderscore d70 & 0.754 & 0.771 \\
Lr\textunderscore 0.001\textunderscore n4\textunderscore d210 & 0.606 & 0.647 \\
Lr\textunderscore 0.001\textunderscore n4\textunderscore d140 & 0.605 & 0.661 \\
Lr\textunderscore 0.001\textunderscore n3\textunderscore d70 & 0.834 & 0.835 \\
Lr\textunderscore 0.001\textunderscore n3\textunderscore d280 & 0.610 & 0.630 \\
Lr\textunderscore 0.001\textunderscore n3\textunderscore d210 & 0.672 & 0.694 \\
Lr\textunderscore 0.001\textunderscore n3\textunderscore d140 & 0.714 & 0.742 \\
Lr\textunderscore 0.001\textunderscore n2\textunderscore d70 & 0.937 & 0.861 \\
Lr\textunderscore 0.001\textunderscore n2\textunderscore d140 & 0.788 & 0.781 \\
Lr\textunderscore 0.001\textunderscore n2\textunderscore d280 & 0.762 & 0.755 \\
Lr\textunderscore 0.001\textunderscore n2\textunderscore d210 & 0.767 & 0.773 \\
Lr\textunderscore 0.0001\textunderscore n4\textunderscore d128 & 0.898 & 0.854 \\
Lr\textunderscore 0.0001\textunderscore n3\textunderscore d70 & 0.894 & 0.855 \\
Lr\textunderscore 0.0001\textunderscore n3\textunderscore d210 & 0.930 & 0.860 \\
Lr\textunderscore 0.0001\textunderscore n3\textunderscore d140 & 0.917 & 0.862 \\
Lr\textunderscore 0.0001\textunderscore n2\textunderscore d70 & 0.912 & 0.864 \\
Lr\textunderscore 0.0001\textunderscore n2\textunderscore d210 & 0.950 & 0.861 \\
Lr\textunderscore 0.0001\textunderscore n2\textunderscore d140 & 0.940 & 0.853 \\
Lr\textunderscore 0.00001\textunderscore n4\textunderscore d70 & 0.707 & 0.726 \\
Lr\textunderscore 0.00001\textunderscore n4\textunderscore d210 & 0.776 & 0.783 \\
Lr\textunderscore 0.00001\textunderscore n4\textunderscore d140 & 0.748 & 0.769 \\
Lr\textunderscore 0.00001\textunderscore n3\textunderscore d70 & 0.735 & 0.758 \\
Lr\textunderscore 0.00001\textunderscore n3\textunderscore d210 & 0.778 & 0.772 \\
Lr\textunderscore 0.00001\textunderscore n3\textunderscore d140 & 0.763 & 0.755 \\
Lr\textunderscore 0.00001\textunderscore n2\textunderscore d70 & 0.713 & 0.749 \\
Lr\textunderscore 0.00001\textunderscore n2\textunderscore d210 & 0.776 & 0.783 \\
Lr\textunderscore 0.00001\textunderscore n2\textunderscore d140 & 0.774 & 0.782 \\
\hline
\end{tabular}
\end{sc}
\end{center}
\end{table}
\clearpage
\section{Math Notations} 

S\textsubscript{t} - state of the molecule at for action t

T - Terminal state/Max length of molecule

S - all possible states

R\textsubscript{t} - reward for the partial molecule at step t

M\textsubscript{t} - candidate molecule

B - target protein

X\textsubscript{c} - Target embeddings from protein-VAE for SELFIES-VAE
G - generator

C\textsubscript{BP} - Critic for Binding Probability prediction

C\textsubscript{SA} - Critic for SA score prediction

C\textsubscript{EA} - Critic for Experimental Affinity prediction

$\Pi(\Theta)$ - optimization policy

N - steps per episode

$P_\Theta$ - probability associated with the action t

$Z_{\text{next}}$ -  ground truth type of the next atom 

$\hat{p}_{\text{type}}^{Z_{\text{next}}}$ - probability that the model assigns to that type at the current step.

$\hat{p}_j^b$ - probability that the model assigns for the distance between $\mathbf{r}_j$ and $\mathbf{r}_{\text{next}}$ to fall into distance bin $b\in B$ at the current step

\textbf{PM-GNN$_{RL}$ model}

$\Pi(\Theta) = \sum_{s_{t} \in S} P_\Theta(s_t)R(s_t) $

$
R(s_t) = 
 \begin{cases}
     \alpha.C_{BP}(B,C_t) + \beta.C_{SA}(C_t),& t = T\\
     0,              & t < T
 \end{cases}
 $

\textbf{3D-MolGNN-$_{RL}$ model}

$\Pi(\Theta) = \sum_{s_{t} \in S} (P_\Theta(s_t) - R(s_t)) $

$P_\Theta(s_t) = P_\Theta(s_t)^{type} + P_\Theta(s_t)^{dist}$

$P_\Theta(s_t)^{type} = -\log \left(\hat{p}_{\text{type}}^{Z_{\text{next}}}\right)$

$P_\Theta(s_t)^{dist} = \sum_{j=1}^{N}\sum_{b\in B} q_j^b \log \left(\hat{p}_j^{b}\right) $ 

$R_1(s_t) = \alpha.C_{BP}(B,C_t) + \beta.C_{EA}(B,C_t) + (1 - \gamma.C_{SA}(C_t))$

$R_2(s_t) = \alpha.C_{BP}(B,C_t) + (1 - \beta.C_{SA}(C_t))$

$R_3(s_t) = \alpha.C_{EA}(B,C_t) + (1 - \beta.C_{SA}(C_t))$


\end{document}